\documentclass[letterpaper, 10 pt, conference]{ieeeconf}  
\usepackage{lipsum}
\usepackage{graphics,graphicx,caption,float,subcaption,booktabs,xcolor,multirow,array,color,ifthen,tabu,colortbl,dblfloatfix,url,xparse,mathtools,patchcmd,algorithm,algorithmic,amssymb,xspace,nicefrac,microtype,amsmath,amsfonts,bm,ragged2e,tikz,stackengine,etoolbox,xpatch,enumerate,xstring,setspace,tabularx,makecell,changepage,cuted,titlesec,wrapfig,tcolorbox, sidecap}

\usepackage[para]{footmisc}
\usepackage{wrapfig}
\usepackage{capt-of}
\usepackage[pagebackref=true,breaklinks=true,colorlinks=true,bookmarks=false,citecolor=blue]{hyperref}

\hypersetup{
colorlinks=true,
linkcolor=blue,
filecolor=magenta,      
urlcolor=blue,
citecolor=blue
}

\usepackage{multicol}
\usepackage[capitalise,noabbrev,nameinlink]{cleveref}
\usepackage[english]{babel}

\makeatletter
\newcommand{\removeParBefore}{\ifvmode\vspace*{-\baselineskip}\setlength{\parskip}{0ex}\fi}
\newcommand{\removeParAfter}{\@ifnextchar\par\@gobble\relax}
\newcommand{\eq}{\begingroup\removeParBefore\endlinechar=32 \eqinner}
\newcommand{\ours}{\texttt{ALAN}\xspace}
\newcommand{\eqinner}[2][aligned]{\endlinechar=32%
\begin{gather}\begin{#1}#2\end{#1}\end{gather}\endgroup\removeParAfter}
\makeatother
\IEEEoverridecommandlockouts                              
\title{\LARGE \bf
ALAN: Autonomously Exploring Robotic Agents \\  in the Real World
}

\author{Russell Mendonca \quad\quad Shikhar Bahl \quad\quad Deepak Pathak \\
Carnegie Mellon University \\
}
\begin{document}

\maketitle
\thispagestyle{empty}
\pagestyle{empty}

\begin{abstract}Robotic agents that operate autonomously in the real world need to continuously explore their environment and learn from the data collected, with minimal human supervision. While it is possible to build agents that can learn in such a manner without supervision, current methods struggle to scale to the real world. Thus, we propose ALAN, an autonomously exploring robotic agent, that can perform tasks in the real world with little training and interaction time. This is enabled by measuring environment change, which reflects object movement and ignores changes in the robot position. We use this metric directly as an environment-centric signal, and also maximize the uncertainty of predicted environment change, which provides agent-centric exploration signal. We evaluate our approach on two different real-world play kitchen settings, enabling a robot to efficiently explore and discover manipulation skills, and perform tasks specified via goal images. Videos can be found at~\url{https://robo-explorer.github.io/}

\end{abstract}

\section{Introduction}

Autonomous robots will need to perform a diverse range of tasks in the real world. Due to the challenges of dealing with uncertainty, deep learning has emerged as a promising approach \cite{levineFDA15,pinto2016supersizing, kalashnikov2018qt} for robotics. A critical challenge for scaling learning based approaches to more complex settings is the task specification problem. Prior works require heavy reward engineering or human demonstrations, which is cumbersome to obtain for performing large numbers of tasks \cite{pastor2009motorskills,ratliff2007imitation,jang2022bc}. This also requires knowledge of the environment, which might be hard to obtain for every domain. Instead, if robots can collect their own data using task-agnostic objectives, they could then autonomously explore their environments and learn interesting skills. 

In the absence of explicit task definitions, the agent should have an efficient way to use all its collected experience for learning. World models \cite{ha2018worldmodels,hafner2019dreamer} provide a means of learning an effective low dimensional representation of raw image observations. Furthermore, if there are certain states where prediction for the world model is difficult, then it likely needs more data for the corresponding part of the environment. This gives rise to a natural intrinsic objective of maximizing model uncertainty \cite{pathak2019disagreement,sekar2020plan2explore, mendonca2021discovering} for exploration. While this does lead to the discovery of interesting behavior, there has been difficulty in scaling such approaches to real world settings since collecting samples on real hardware is very time-intensive. We ask if there is a different task-agnostic objective that can enable robots to \textit{more efficiently} explore? 

\begin{figure}[h!]
    \centering
    \includegraphics[width=\linewidth]{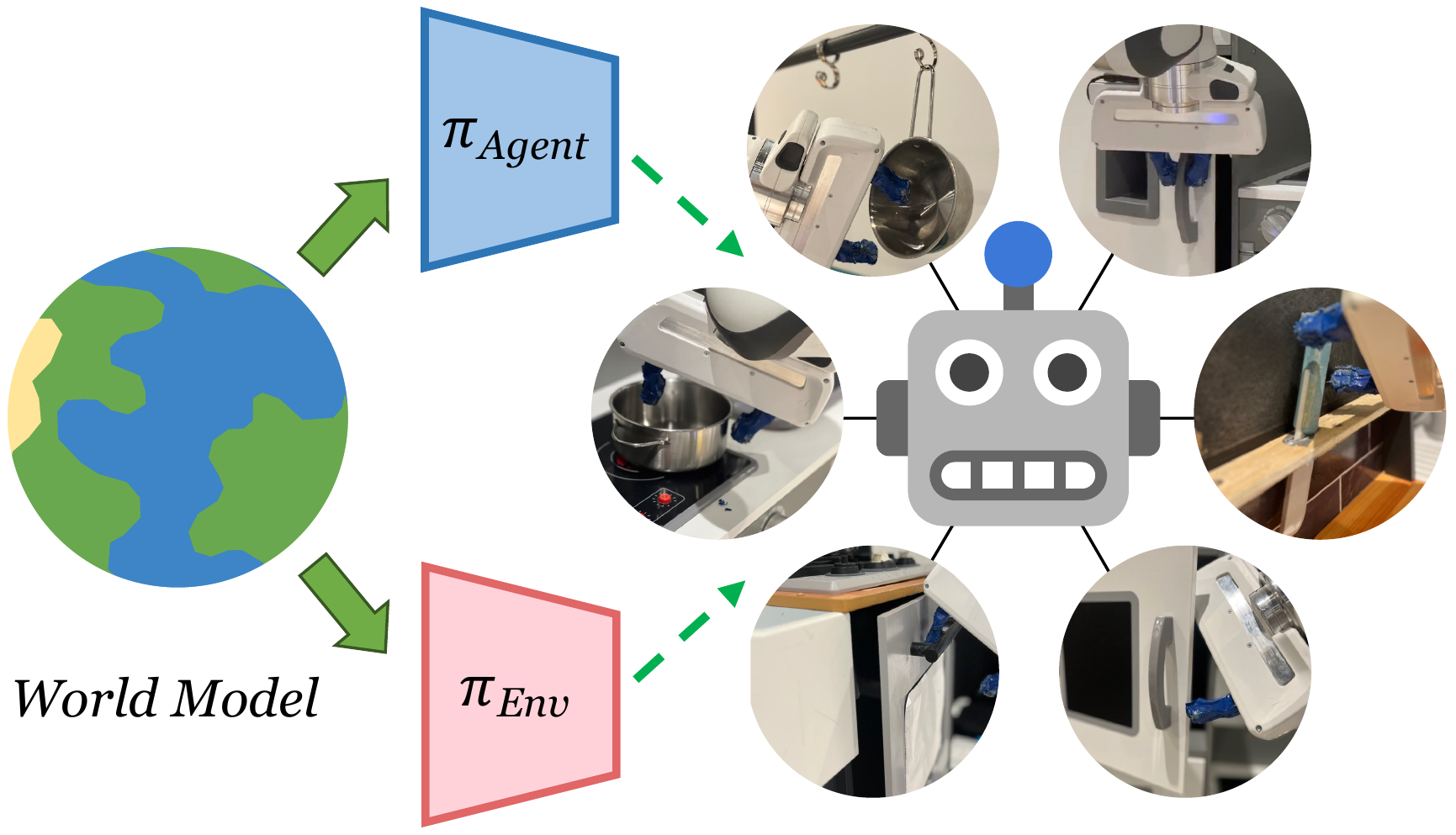}
    \caption{\small We present ALAN, an approach for real world robotic exploration in challenging manipulation environments.}
    \label{fig:teaser}
    \vspace{-0.2in}
\end{figure}

In order to address the above question, we present \ours, an efficient autonomous real robot explorer. Our key insight is that interesting behavior for robots in the manipulation setting mostly involve \emph{interactions with objects, which cause changes in the visual features of the observations}. Thus, seeking to maximize the change in these visual features can be a useful objective for robots to optimize. Furthermore, if agents learned to model the change in the environment, they can take actions to maximize uncertainty in the \emph{object space} of the environment, as opposed to the full space consisting of both the robot body and the surrounding environment. Seeking to maximize information related to objects in the environment will lead to much more efficient exploration, since the robot will prioritize actions that lead to richer contact interactions. We note that maximizing model uncertainty, (whether in the object space or full image space) is `agent-centric', since it is dependent on the agent's belief, as opposed to  simply maximizing the environment change which is `environment centric'. The latter is a constant signal agnostic of the agent's internal mental model. We show that leveraging both these objectives can enable a real robot to effectively explore multiple challenging real-world environments, and then perform tasks of interest.

\begin{figure*}[t!]
    \centering
    \includegraphics[width=\textwidth]{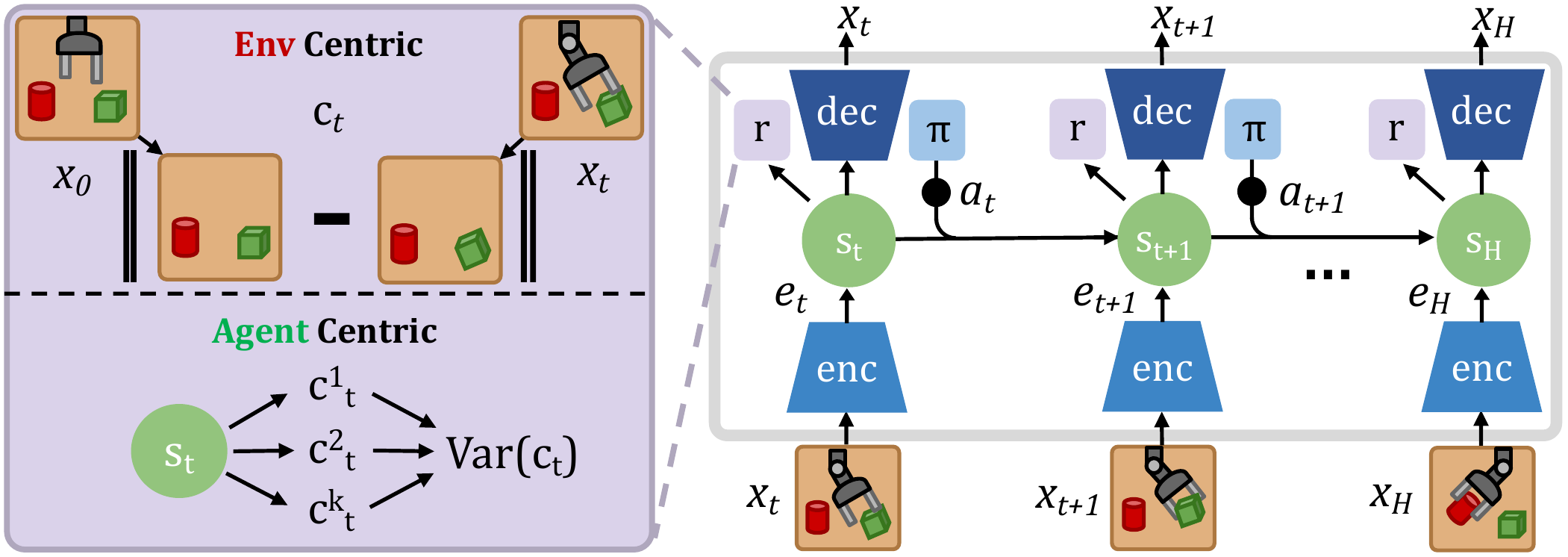}
    \vspace{-0.1in}
    \caption{\small We propose Autonomous Learning Agents (\ours) that can enable robots to collect rich data from their environment efficiently. The agent utilizes environment change, both directly as an environment-centric signal, as well as 
    modelling the change and taking actions that maximize uncertainty in change space, which provides agent-centric signal.}
    \label{fig:achiever-success-plots}
    \vspace{-0.1in}
\end{figure*}

The main contribution of this work is \ours, an efficient real world exploration algorithm, that seeks to take actions that maximize change in the environment, and maximize uncertainty about its internal model of how changes occur in the environment. This approach encourages the robot to interact with objects, and hence collect data relevant to learning manipulation skills faster. We show that our approach enables a Franka Emika robot to effectively explore without any supervision signal in two different, challenging play-kitchen environments using less than 150 interaction trajectories. The robot can then perform user-specified tasks via goal images in a zero-shot manner, including picking up a knife, and opening a cabinet, fridge or shelf.

\section{Related Work}

\noindent\textbf{Exploration$\quad$} In reinforcement learning (RL), exploration has been studied in various contexts ranging from tabular settings to high-dimensional continuous spaces. For simple discrete settings, analysis of exploration has included state visitation counts \cite{strehl08} and probability distributions over visited states \cite{duff2002bayesadaptive, poupart2006analytic}. For high-dimensional input spaces such as images, previous works have used neural networks to approximate state counts \cite{bellemare2016cts, ostrovski2017count,tang2017exploration} and for sampling goals \cite{zhang2020automatic, pong2019skewfit}. Another approach to describe intrinsic reward for exploration is to use either the error \cite{pathakICMl17curiosity} or uncertainty \cite{lowrey2018polo, osband2018randomprior} in prediction about how the environment and agent would interact. Pathak et al. \cite{pathak2019disagreement} proposes a differentiable intrinsic reward which measures disagreement using the variance of the prediction of an ensemble of models. Sekar et al. \cite{sekar2020plan2explore} leverages a similar disagreement-based intrinsic reward, but explores in the imagination space of a learned world model \cite{hafner2018planet, hafner2019dreamer}. 

\noindent\textbf{Autonomous Learning in the Real World$\quad$} Training agents in the real world is challenging for a host of reasons, and one of these is the difficulty of providing supervision to the agent. Some prior approaches have designed task specific rewards~\cite{levine2013guided,levineFDA15}. However, it is infeasible to define all of the tasks that are possible for the robot to perform, and further there is no guarantee that the designed rewards will allow for the task to be solved efficiently and robustly. There are a number of approaches that provide self-supervision for agents based on mutual information objectives \cite{diayn, dads, discern}, which enables the learning of skill-spaces. However, many of these learned skills are not semantically different and have been difficult to apply to real-world manipulation. Other approaches involve selecting goals from experience. This can directly come from previously seen states \cite{andrychowicz2017hindsight}, from a generative model \cite{ nair2018visual, pong2019skew, zhang2020automatic}, or from the imagination space of a world-model 
\cite{mendonca2021discovering}. While these approaches have shown better results for real-world manipulation, they are still limited in scope, since they require lots of samples for learning. A key reason is that it is difficult for the robot to know \textit{what} to focus on while exploring. Efforts have been made to initialize such approaches from priors of human behavior, such as from internet data \cite{shao2021concept2robot, bahl2022human, chen2021dvd}, however, such methods are not able to learn in an autonomous fashion. Our approach provides an effective new metric that enables efficient self-supervision, and also leverages visual priors to focus on parts of the scene that are more interesting for exploration and discovery of useful skills.

\section{Background}

\noindent\textbf{Model-Based RL and Planning$\quad$} A Markov Decision Process (MDP) is defined by a set of states  $\mathcal{S}$, actions $\mathcal{A}$, transition probabilities between states conditioned on actions, $\mathcal{T}(s_{t+1}|s_t,a_t)$, a initial state distribution $\mathcal{S}_0$, a reward function  $\mathcal{R}(s_t, a_t)$. The goal of a model based RL algorithm is to learn a function $f_\theta(s_{t+1}|s_t,a_t)$ which best approximates the the true transition dynamics $\mathcal{T}$ of the MDP. While planning, the Cross-Entropy Method (CEM) can be used to find the best set of actions $a_{1:T}$, which produce the highest reward under the trained dynamics model $f_\theta$.

\noindent\textbf{Intrinsic Motivation$\quad$} When learning a dynamics model of the world, $f_\theta(s_{t + 1} | s_t, a_t)$, it is possible to use the quality of the model as an intrinsic reward. For instance, Pathak et al. \cite{pathakICMl17curiosity} use model prediction error as reward $$r_t = || f_\theta(s_{t + 1} | s_t, a_t) - s_{t + 1}||$$
However, this formulation is dependent on environment dynamics, and thus needs a policy-gradient approach to optimize it, since future states need to be observed before this metric can be computed. Instead, \cite{pathak2019disagreement} proposes to minimize the \textit{disagreement} between an ensemble of dynamics model $f_{\theta^{(k)}}$ for $k = 1, ..., M$, which is a fully differentiable objective in terms of the current state and action, which we utilize in our work. The disagreement reward can be described as:
$$\mathbb{E}_{s_t, a_t, s_{t+1} \sim \rho(s)}[\text{Var}_k({f_{\theta^{(k)}}})]$$

\section{Autonomous Real World Robot Learning}

\begin{figure}[t!]
    \centering
    \begin{subfigure}[b]{0.49\linewidth}
    \includegraphics[width=\linewidth]{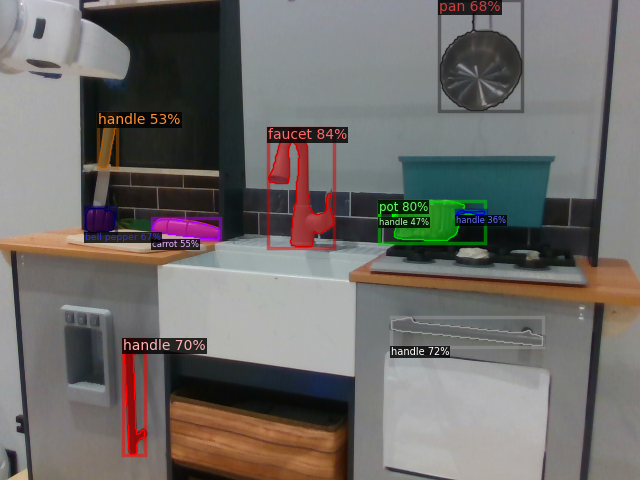}
    \end{subfigure}
    \begin{subfigure}[b]{0.49\linewidth}
    \includegraphics[width=\linewidth]{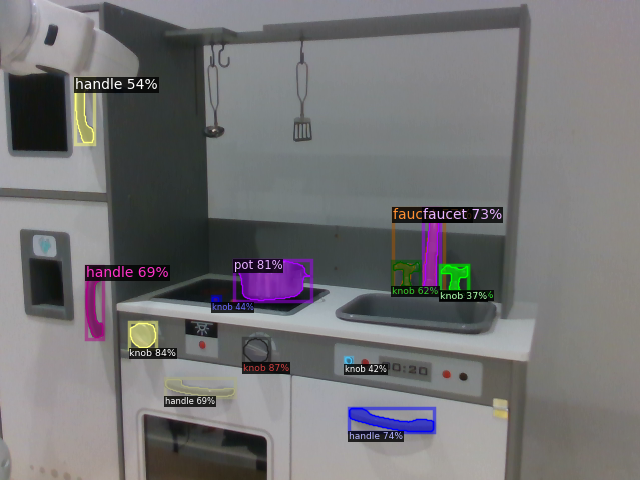}
    \end{subfigure}
    \caption{ \small Visualizations of the object detections, using \cite{detic}. The masks selected to study exploration are the knife, pan and rightcabinet handle from kitchen1 (left), and the topshelf, fridge handles and pot from kitchen2 (right).}
    \label{fig:detic}

\end{figure}

Intelligent agents should be able to perform diverse tasks in complex, real world environments. There are three major challenges to this:  (1) There is a large space of possible interactions, especially in continuous control. (2) It is difficult to obtain any reward signal without human supervision. (3) There is a large cost for collecting data with real hardware.

To this end, we propose \ours, an autonomous robot learning algorithm that is able to efficiently explore in the real world, and learn useful manipulation skills for various objects. \ours defines a novel intrinsic exploration objective for the agent to direct its behavior. This novel objective has an environment-centric component and an agent-centric component. Moreover, we use offline visual data to reduce the search space for the robot, by identifying the locations of potential interesting and complex interactions for the robot.

\subsection{World Model}
The robot observations consist of a stream of high-dimensional raw RGB images. These can be effectively processed using world models \cite{ha2018worldmodels, schmidhuber_curiosity, schmidhuber1991curiousmodel}, which learn compact low-dimensional latent spaces that contain temporal information and enable efficient forward prediction. We use the Recurrent state-space model (RSSM), from \cite{hafner2018planet, hafner2019dreamer, hafner2020dreamerv2}, which learns latent features with deterministic and stochastic components to model long-range dependencies and uncertainty in the environment respectively. Specifically, the world model has the following networks:

\eq{
&\text{Image Encoder} && h_t = \text{enc}_{\theta}(x_{t}) \\
&\text{Dynamics Prior} && p_\theta(s_{t + 1}| s_{t}, a_{t})\\
&\text{Image Decoder} && f_\theta(x_t| s_t) \\
&\text{Dynamics Posterior} && q_\theta(s_{t + 1} | s_t, a_t, h_{t + 1})\\
&\text{Embed Decoder} && g_\theta(e_t| s_t) \\
}

The latent features are trained to reconstruct image observations, while also preserving dynamics information using variational inference and the ELBO loss \cite{rezende2014vae, kingma2013vae}. 
In addition to providing useful representations for control, world models also provides a means for agents to drive their own behavior in the absence of supervision. This involves taking actions that maximize the uncertainty of model predictions \cite{pathak2019disagreement, sekar2020plan2explore, mendonca2021discovering}, leading to information gain for the agent. Since this is dependent on the agent's internal belief, we call this kind of exploration `agent-centric'. In the next section we first consider a different source of signal which is environment-centric, and then discuss how it can be used to augment agent-centric exploration as well. 

\subsection{Environment Change}
Seeing as how interesting manipulation behavior often involves changes in object states, and how this corresponds to change in visual features, we seek to autonomously estimate environment change from observed data. To capture environment interaction, the change metric should ignore differences in the robot's position, and only highlight movement of objects in the scene \cite{bahl2022human}. How then can we extract these ground truth change images from incoming image observations?  

Our source of signal is assuming knowledge of the visual appearance of the robot, using which we train a segmentation model $m_\phi(.)$ to mask out the robot from the scene. Training this model is a one time cost, since the robot appearance is invariant across multiple tasks in the environment and even across different domains. We can use this model to measure the environment change $f_c$ between an image pair $x_i$, $x_j$:

\begin{equation}
\begin{aligned}
    f_c(x_i, x_j) = f(||m_\phi(x_i) - m_\phi(x_j)||_2 ,  \\
    || \Psi(m_\phi (x_i)) - \Psi( m_\phi (x_j))||_2)
\end{aligned}
\label{eq:ec_rew}
\end{equation}

Here the heuristic function $f$ takes into account pixel distance, blurring to remove shadows and reflective surface artifacts, and $\Psi$ denotes visual features from a pretrained segmentation network \cite{he2017mask}, and returns a binary image indicating the pixels where change has been detected. We further apply a threshold for the change image, in order to minimize false detections. We don't require this change function to be fully accurate, and have found that our approach is robust to some error in the change image. For an image $x_t$ from a trajectory $\mathcal{T}$, the corresponding change $c_t$ can be defined as $f_c(x_t, x_{t-1})$ or $f_c(x_t, x_0)$, where $x_0$ is the first image in $\mathcal{T}$. We found the latter produced better exploration, likely because the change between consecutive image frames is very small and is diffuclt to reliably detect.

\begin{algorithm}[t!]
\caption{ \ours  : Exploration}
\label{alg:method}
\begin{algorithmic}[1]
\REQUIRE Robot segmentation model $m_\phi$
\REQUIRE Off-policy RL algorithm $\mathcal{A}$
\REQUIRE Visual Priors (\ref{subsec:vis_priors}) for structured space \\
\hspace{-0.28in} \textbf{Initialize: }World Model $\mathcal{W}$, Biasing policy $\pi$, Dataset $R_D$

\WHILE{Sampling}
\STATE Run $\pi$ through $\mathcal{W}$ in imagination to obtain $\{ \hat{a}_t \}_H$
\STATE Run CEM with $\mathcal{W}$ using objectives \ref{eq:max_ec} and \ref{eq:disag_rew}, and $\{ \hat{a}_t \}_H$ as initial proposals, to collect trajectory $\mathcal{T}$
\STATE Label $\mathcal{T}$ with $c_t = f_c(x_t, x_0)$ (Eq. \ref{eq:ec_rew}), add to $R_D$
\ENDWHILE
\WHILE{Training}
\STATE $\mathcal{S_D}$ = Top $N_A$ trajs in $\mathcal{R_D}$, based on $\sum{c_t}$
\STATE Update $\pi$ using $\mathcal{A}$ on $\mathcal{S_D}$
\STATE Update $\mathcal{W}$ using $R_D$
\ENDWHILE
\end{algorithmic}
\label{algo:expl_method}

\end{algorithm}

\noindent\textbf{Environment-centric exploration$\quad$} Using the norm of the change image as a metric, we can use off-policy RL \cite{peng2019advantage, nair2020awac, iql} approaches to train a policy for control. The approach we use is to incorporate the metric into a world model by training the features $s_t$ to also predict the change in the environment between observation $o_t$ and the initial observation of the trajectory $o_0$, by adding an additional change predictor module $r_\theta(c_t | s_t)$. This is optimized by maximizing $\mathbb{E}[\log p(c_t|s_t)]$, similar to the image decoder, where $c_t$ is the change image. While exploring under this objective, we optimize:

\begin{equation}
\label{eq:max_ec}
   \arg \max_{a_1..a_T} \mathbb{E}_{s \sim \rho(s)}[\sum (r_\theta(c_{t + 1}|s_{t+1}) \big|  s_t , a_t)] 
\end{equation}

\noindent\textbf{Change-space agent-centric exploration$\quad$} Since the agent now models the environment change in its internal belief, it can leverage errors in this model to direct exploration. Just as previous exploration approaches maximize uncertainty of next state using the model \cite{pathak2019disagreement, sekar2020plan2explore} the agent can maximize uncertainty over the \emph{change} prediction. Thus, the agent will collect data that leads to information gain specifically about how the objects in the environment move, avoiding being stuck gathering information pertaining to the robot's own body. Thus the agent will collect data that includes more information about object interactions. Specifically, we implement this by training an ensemble of models for $p(c_{t+1} | c_t, a_t )$, where $c_t$ and $a_t$ are the predicted change and action at time t respectively. To maximize uncertainty in change space, we optimize for actions that maximize the variance of the ensemble prediction (here $s_t$ is a latent sampled from the world model) :  

\begin{equation}
\label{eq:disag_rew}
    \arg \max_{a_1..a_T} \mathbb{E}_{s \sim \rho(s)}[\text{Var}_k({p_{\psi^{(k)}}}(r_\theta(c_{t + 1}|s_{t+1})) \big|  r_\theta(c_t|s_t) , a_t)] 
\end{equation}

\noindent\textbf{Control$\quad$} Now that the features of the world model are trained to predict environment change, we can explore by planning through the model adding the objectives from \ref{eq:disag_rew} and \ref{eq:max_ec}. We use the Cross entropy method \cite{rubinstein1997cem} for planning, where we sample action proposals from an initial distribution, pick the top trajectories based on reward and refit the sampling distribution. Further, we train Advantage Weighted Regression (AWR) on the collected offline trajectories to maximize the environment change in the feature space of the world model. When sampling, given an observation, we first run the learned AWR policy through the model in imagination to get a sequence of actions. We use this as the mean of the initial normal sampling distribution for CEM, to bias the optimization procedure towards trajectories that are likely to have high environment change. We summarize the full exploration method in Alg. \ref{alg:method}, including both sampling and training which are  run asynchronously.

\subsection{Leveraging Visual Priors}
\label{subsec:vis_priors}
While environment change and ensemble disagreement can provide useful signal for driving behavior, the large work spaces in the real world pose a major challenge for robots. Exploration methods often spend a lot of time in free space, and collect a large number of samples without interacting with any objects. This is undesirable since this data contributes little to learning manipulation skills. Our approach to avoiding this is to leverage visual priors from offline data, helping understand \textit{what} to explore. One instance of this is to leverage object-detectors to initialize the robot near regions of interest. Recent models \cite{detic} are quite robust and can identify objects even in cluttered scenes. Using RGBD cameras and homography calibration for the robot with the cameras, we can then initialize the robot end effector close to the center of the object point-cloud, thus ensuring that data-collection is more likely to see object interactions. This approach does not preclude training on undetected objects, since the robot can always randomly sample points in the full workspace to initialize at later, and will likely be more proficient after it has learned skills efficiently on all the detected objects. 
For a image that has $k$ detected masks  $M_1, ..., M_k$, the robot can arbitrarily pick any mask for initialization every episode. However, in order to study exploration for independent objects separately, we enforce that the robot needs to reset to the same mask each time, and since this choice can be arbitrary, we also specify which mask should be selected, so that different methods can be evaluated on the same objects. We use the same visual prior for the baselines and ablations to make the exploration space feasible.

\begin{figure}
    \centering
    \includegraphics[width=\linewidth]{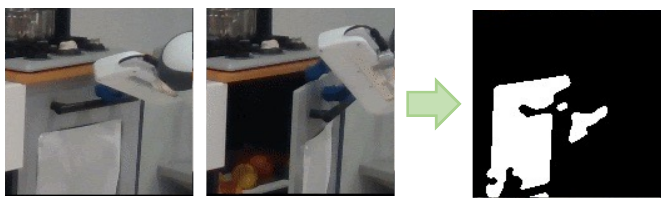}
    \caption{ \small An example of the change image extracted from a pair of images, as described in Equation \ref{eq:ec_rew}. This is a binary image that detects pixels where change has occured. }
    \label{fig:ec}
\end{figure}

\subsection{Achieving goals}
\label{sec:achiever}
Given the contact-rich data collected by the exploration controllers, how can we use this data to perform useful tasks? It is possible for the agent to sample goals from previously seen exploration data. Since the agent sees interesting data, any possible state can be a goal. Concretely, given some human sampled goal images, $x_g$, we leverage recent advances in goal-conditioned imitation learning, especially methods that leverage Nearest Neighbor-based techniques in a self-supervised representation space \cite{pari2021surprising}. Our policy, $\pi_{knn}$ scans through image features \cite{nair2022r3m} in the exploratory data, and selects the top trajectory matches:

\begin{equation}
\label{eq:gc-policy}
    \tau^\star =  \text{argmin}_i \min_{x_j \in \tau_i} ||\phi(x_g) - \phi(x_j)||_2
\end{equation}

 Since our method has seen interesting trajectories, it is more likely to see semantically useful goals, and thus when a human provided goal $x_{gh}$ is given, more likely to reach it.

\begin{figure}[t!]
    \centering
    \includegraphics[width=0.8\linewidth]{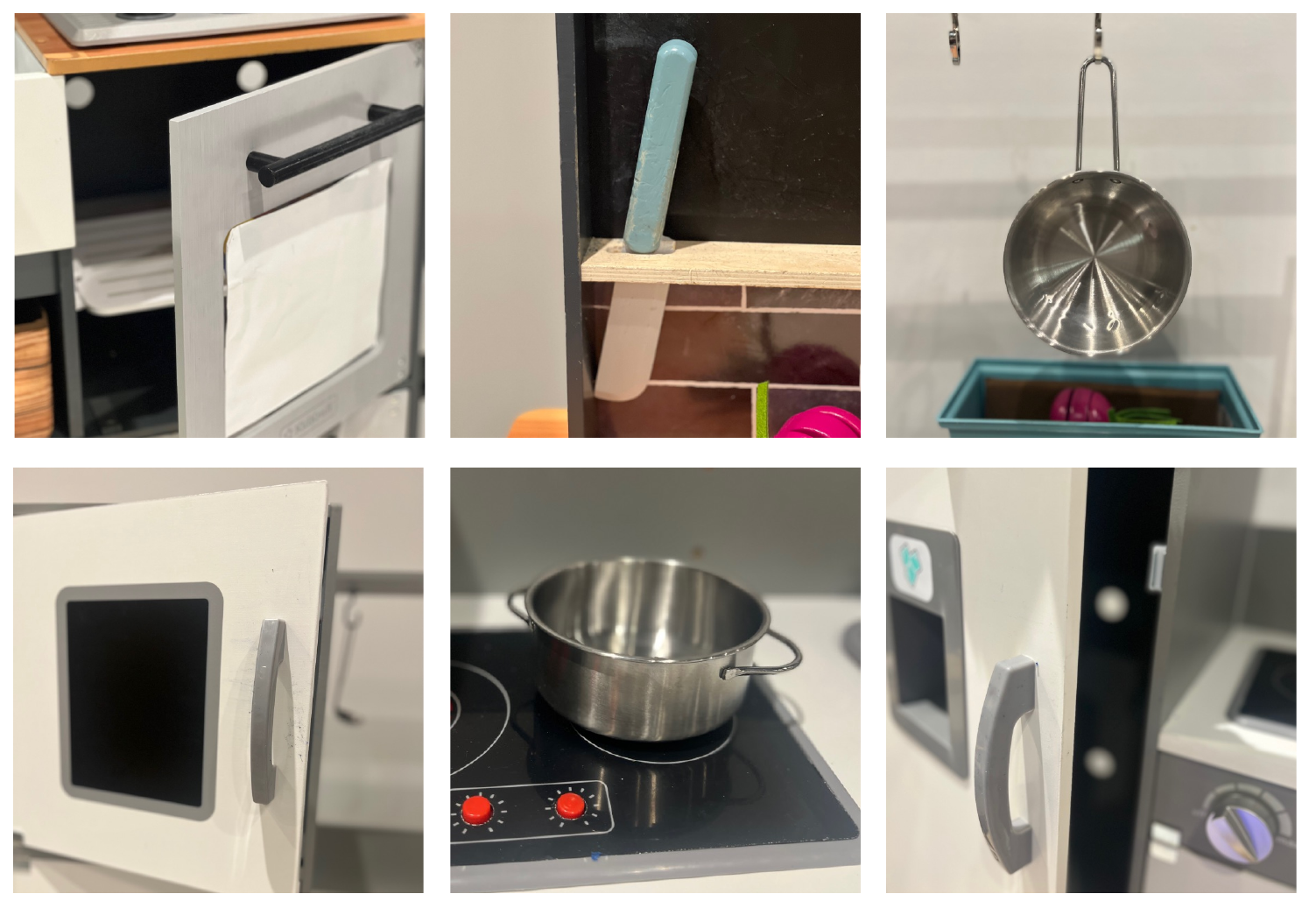}
    \caption{\small We explore on 6 settings across two play kitchens. Top, from left: cabinet, knife, pan (kitchen1). Bottom, from left: top shelf, pot, fridge (kitchen2).}
    \label{fig:tasks}
    \vspace{-0.1in}
\end{figure}

\section{Experimental Setup}

In our experiments, we ask the following questions : 1) Does our system enable autonomous exploration and discovery of interesting states in complex real world environments? 2) How does the quality of this data compare to that of current SOTA approaches? 3) Is it possible to use this data to reach human specified goals to perform useful tasks?

\noindent\textbf{Real World Setup$\quad$} We tested our system on a Franka Panda 7-DOF robot, and on two different real-world kitchen play-sets, which have many diverse objects and possible manipulation tasks, comprising a very large search space (both are about 100cm X 100cm X 100cm). Specifically, we investigate 6 object regions across two kitchens detected by our visual prior approach \cite{detic}, as shown in Figure~\ref{fig:detic}. Namely, these are the knife, cabinet and the hanging pan from the first kitchen, and the top shelf, fridge and pot from the second kitchen (Figure~\ref{fig:tasks}). During training we provide minimal resets via human intervention, and only when the object is in an un-resettable state (for example when the knife or pan has fallen down), or for safety reasons. Our setup uses 2 cameras to cover the entire scene, and the observation space consists of a single 128X128 size RGB image from the camera that is farther from the robot end effector, which provides a more complete view of interaction. We execute 6-DOF control on the arm along with open-close gripper action. The change image is resized to a 32X32 binary image for prediction. Training and sampling are run asynchronously.

\begin{figure}[t!]
    \centering
    \begin{subfigure}[b]{0.49\linewidth}
    \includegraphics[width=\linewidth]{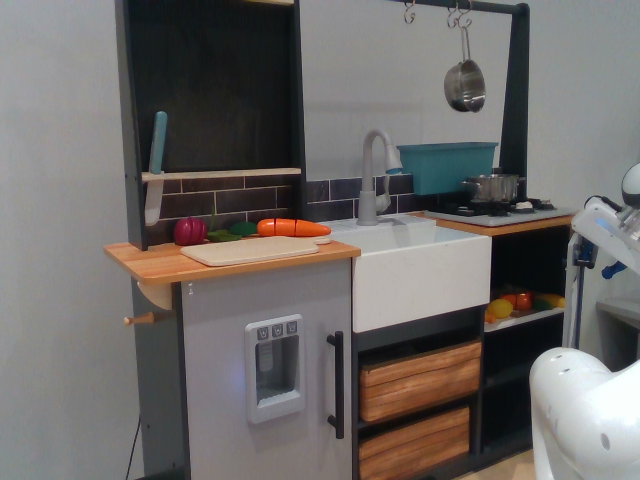}
    \caption{\small Cabinet}
    \end{subfigure}
    \begin{subfigure}[b]{0.49\linewidth}
    \includegraphics[width=\linewidth]{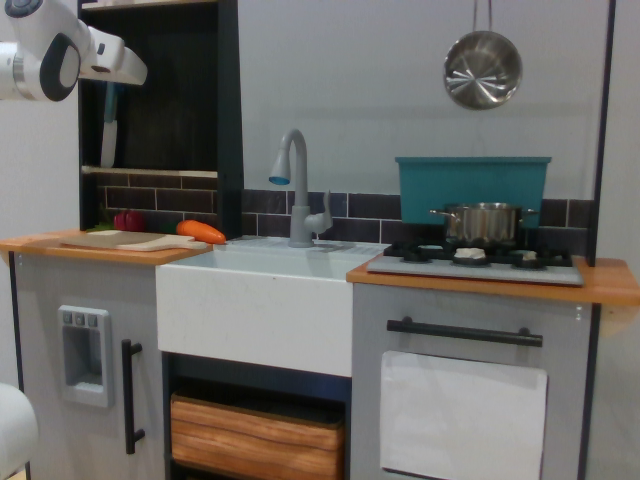}
    \caption{\small Knife}
    \end{subfigure}
   
    \begin{subfigure}[b]{0.49\linewidth}
    \includegraphics[width=\linewidth]{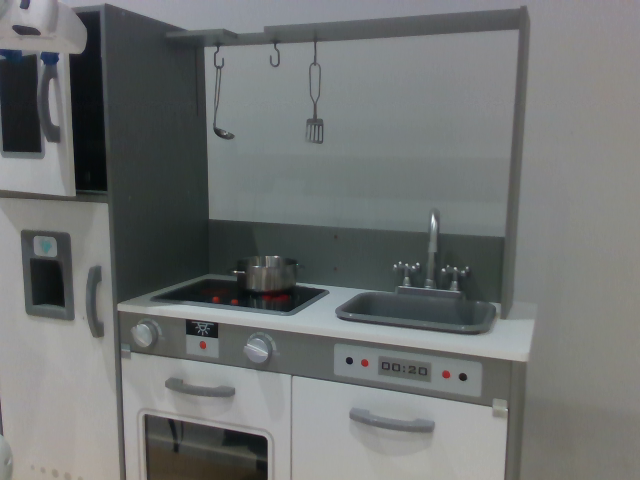}
    \caption{\small Top Shelf}
    \end{subfigure}
    \begin{subfigure}[b]{0.49\linewidth}
    \includegraphics[width=\linewidth]{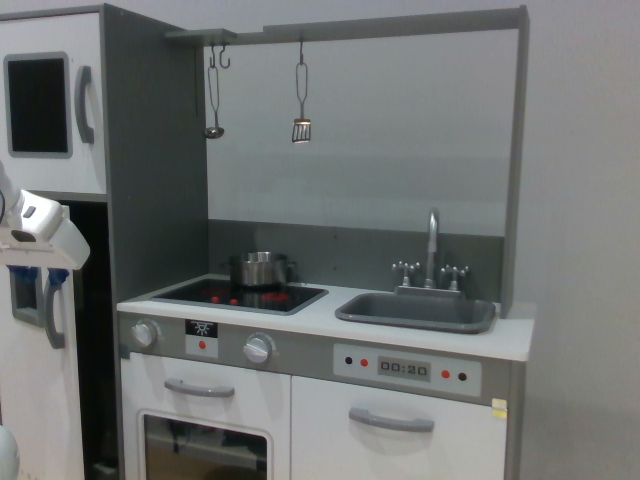}
    \caption{\small Fridge}
    \end{subfigure}
   
    \caption{\small Manually specified goals used for zero-shot evaluation, after the completion of the exploration phase. }
    \vspace{-0.25in}
    \label{fig:goal-imgs}
\end{figure}

\noindent\textbf{Training Procedure$\quad$} For each of the regions, we first collect a random dataset of 25 trajectories. All collected trajectories are 20 timesteps long. The world models in all methods use an RSSM \cite{hafner2018planet}, and the image encoders and decoders use the NVAE architecture \cite{vahdat2020nvae}. To extract the environment centric metric, we train a Mask RCNN model \cite{he2017mask} on 200 images using data from both play kitchens. 

\noindent\textbf{Baselines and Ablations$\quad$} We compare against \texttt{LEXA}\cite{mendonca2021discovering}, a state-of-the art self-supervised exploration approach for continuous control in manipulation settings. \texttt{LEXA} outperforms various other self-supervised approaches, \cite{pong2019skew, diayn, ddl} on a complex simulated kitchen environment both in terms of the exploratory data seen, and the success rate of reaching discovered goal images. We provide this baseline with the same world model architecture as \ours. 
 
Next, we ablate the need of our agent-centric module, which explores in the change space. This is to test our hypothesis that the robot should continually collect data where the model predictions regarding environment change are inaccurate. We test if this ability is crucial, by running the environment-centric exploration model, which only uses the intrinsic reward described in Equation~\ref{eq:ec_rew}. We run two versions of this, \texttt{EC} which uses the model for planning, and \texttt{AWR} which just uses the trained AWR policy, without planning.

\begin{figure*}[t!]
    \centering
    \begin{subfigure}[b]{0.325\linewidth}
    \includegraphics[width=\linewidth]{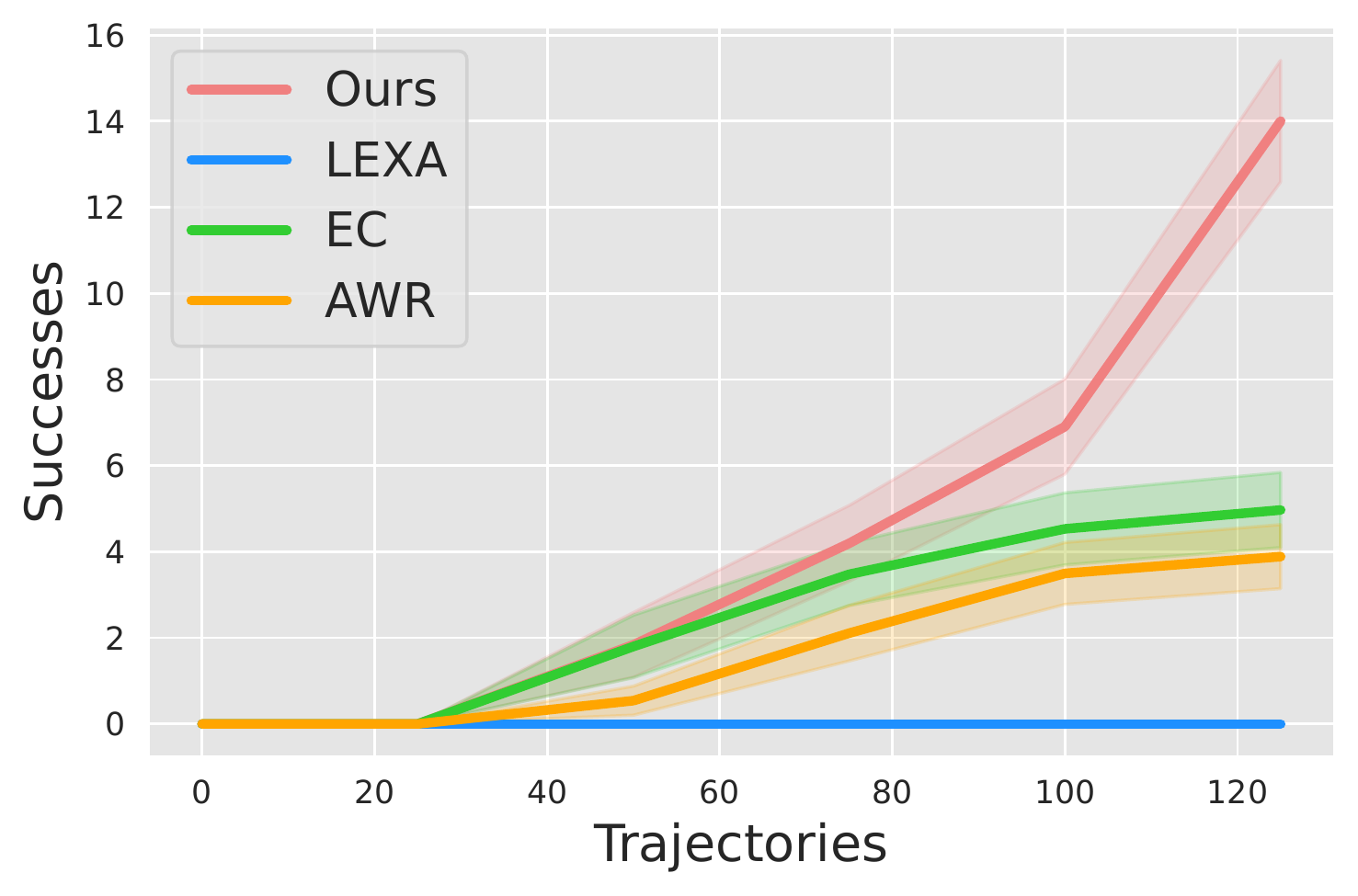}
  
    \caption{\small Knife}
    \end{subfigure}
    \begin{subfigure}[b]{0.325\linewidth}
    \includegraphics[width=\linewidth]{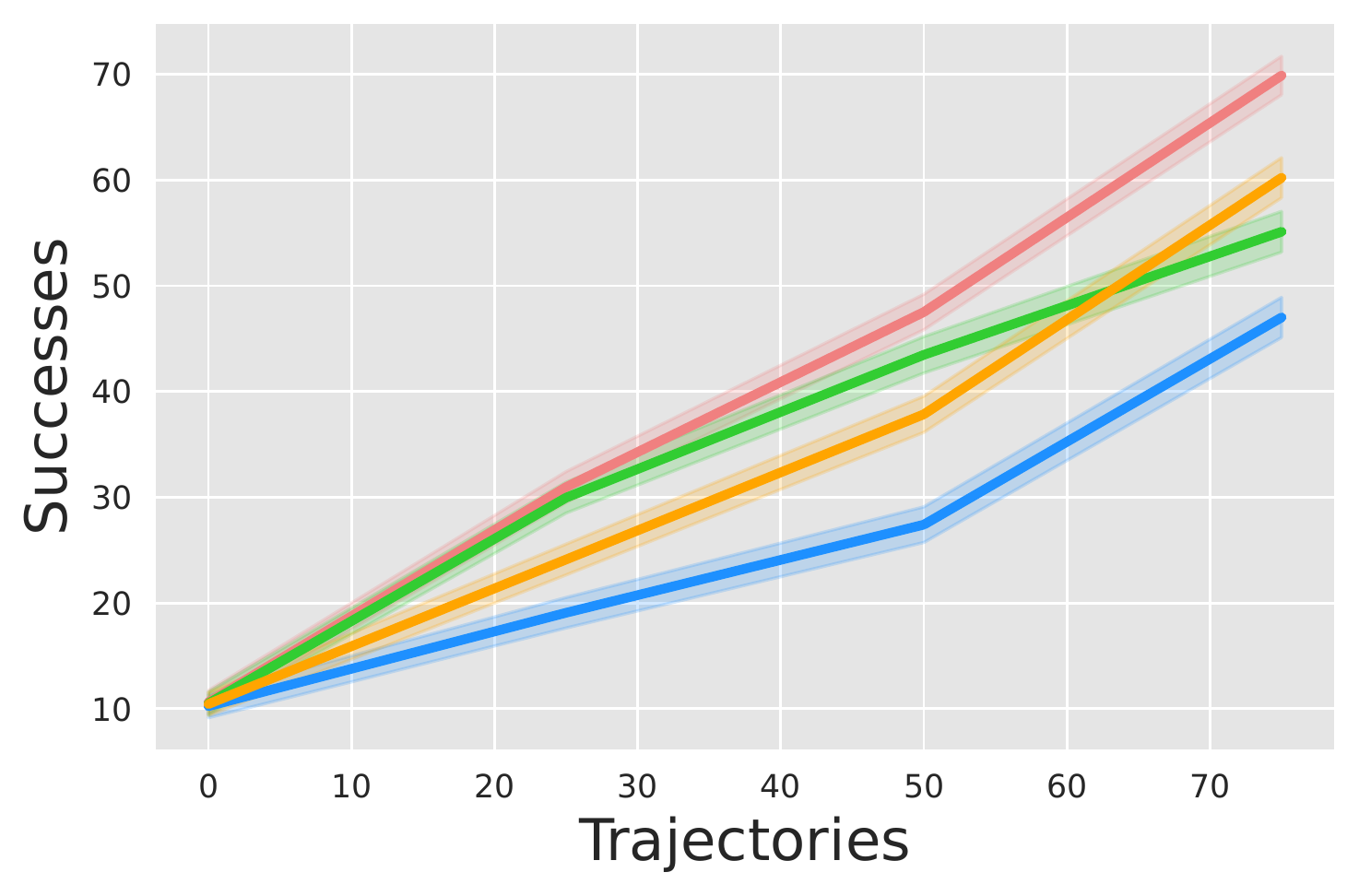}
   
    \caption{\small Cabinet}
    \end{subfigure}
    \begin{subfigure}[b]{0.325\linewidth}
    \includegraphics[width=\linewidth]{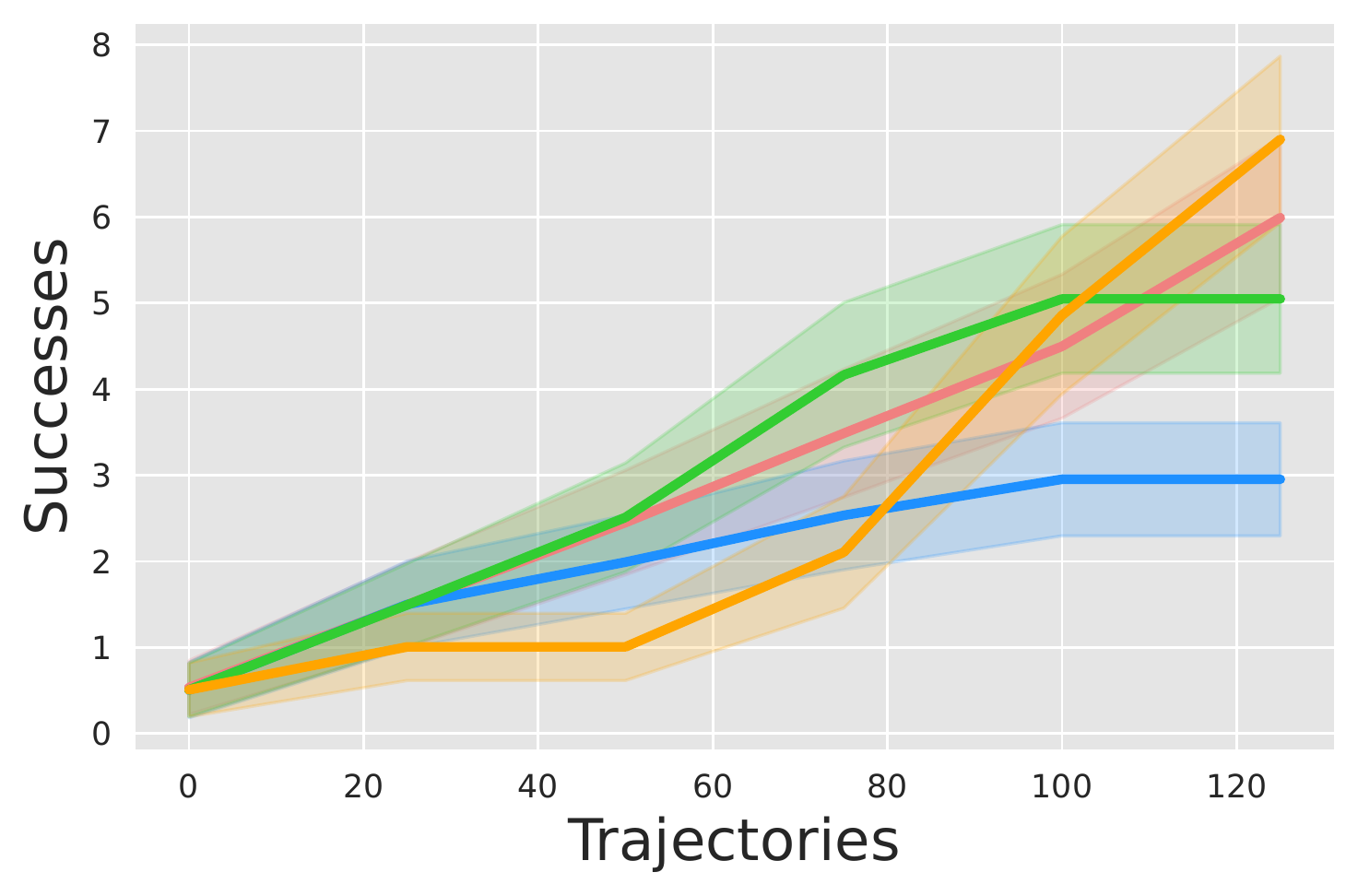}

    \caption{\small Pan}
    \end{subfigure}
    \begin{subfigure}[b]{0.325\linewidth}
    \includegraphics[width=\linewidth]{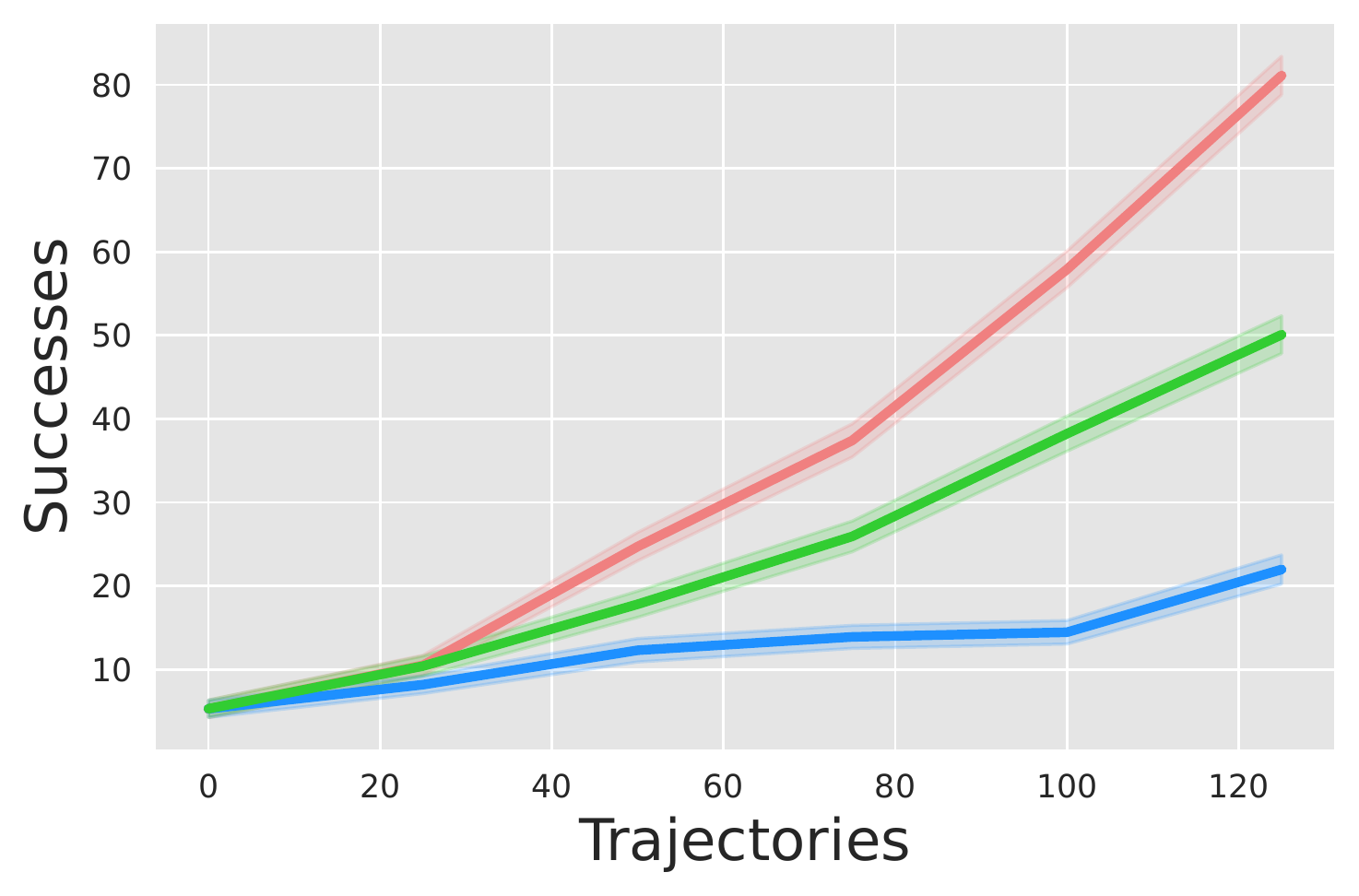}

    \caption{\small Fridge}
    \end{subfigure}
    \begin{subfigure}[b]{0.325\linewidth}
    \includegraphics[width=\linewidth]{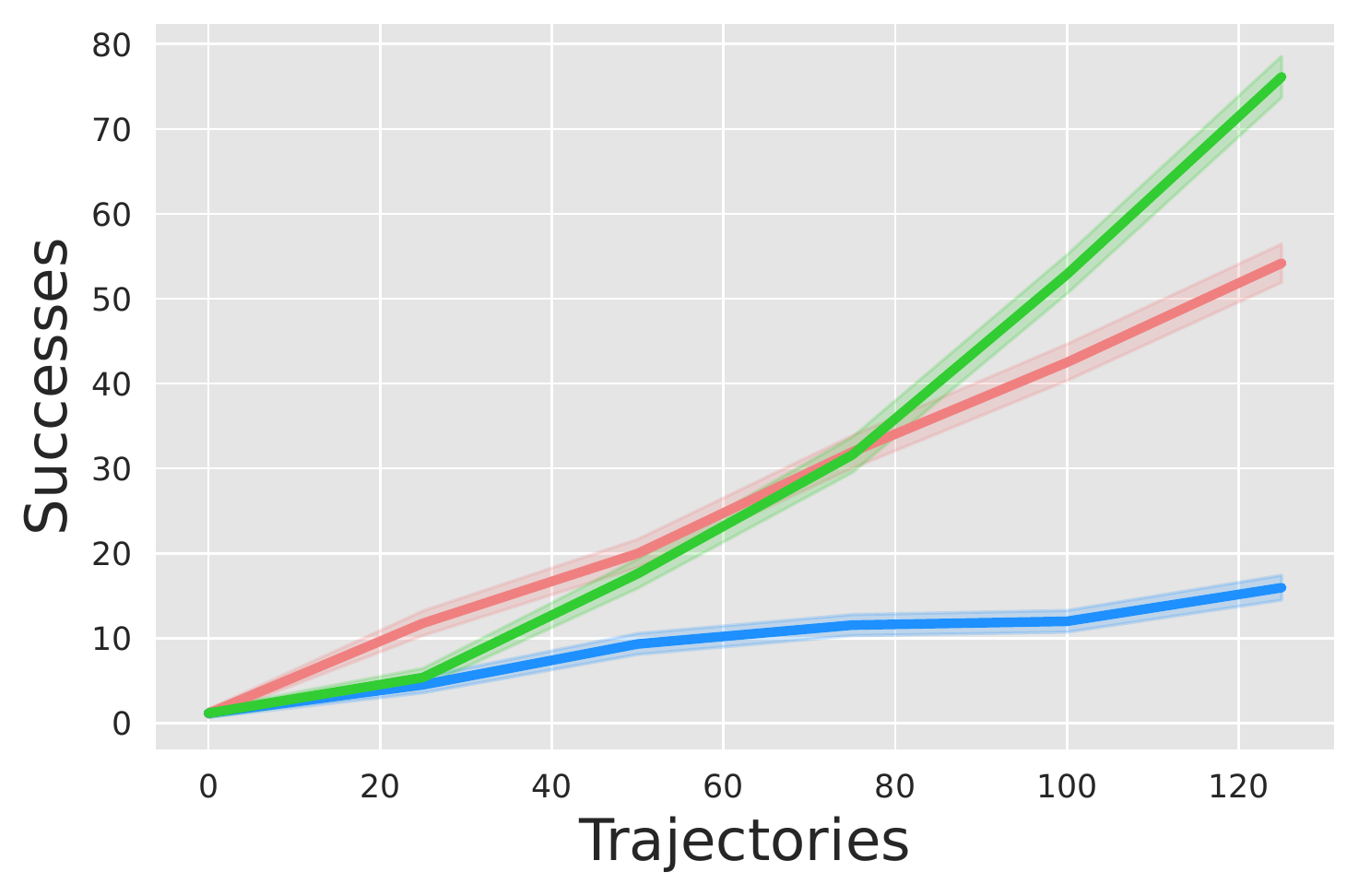}

    \caption{\small Top Shelf}
    \end{subfigure}
    \begin{subfigure}[b]{0.325\linewidth}
    \includegraphics[width=\linewidth]{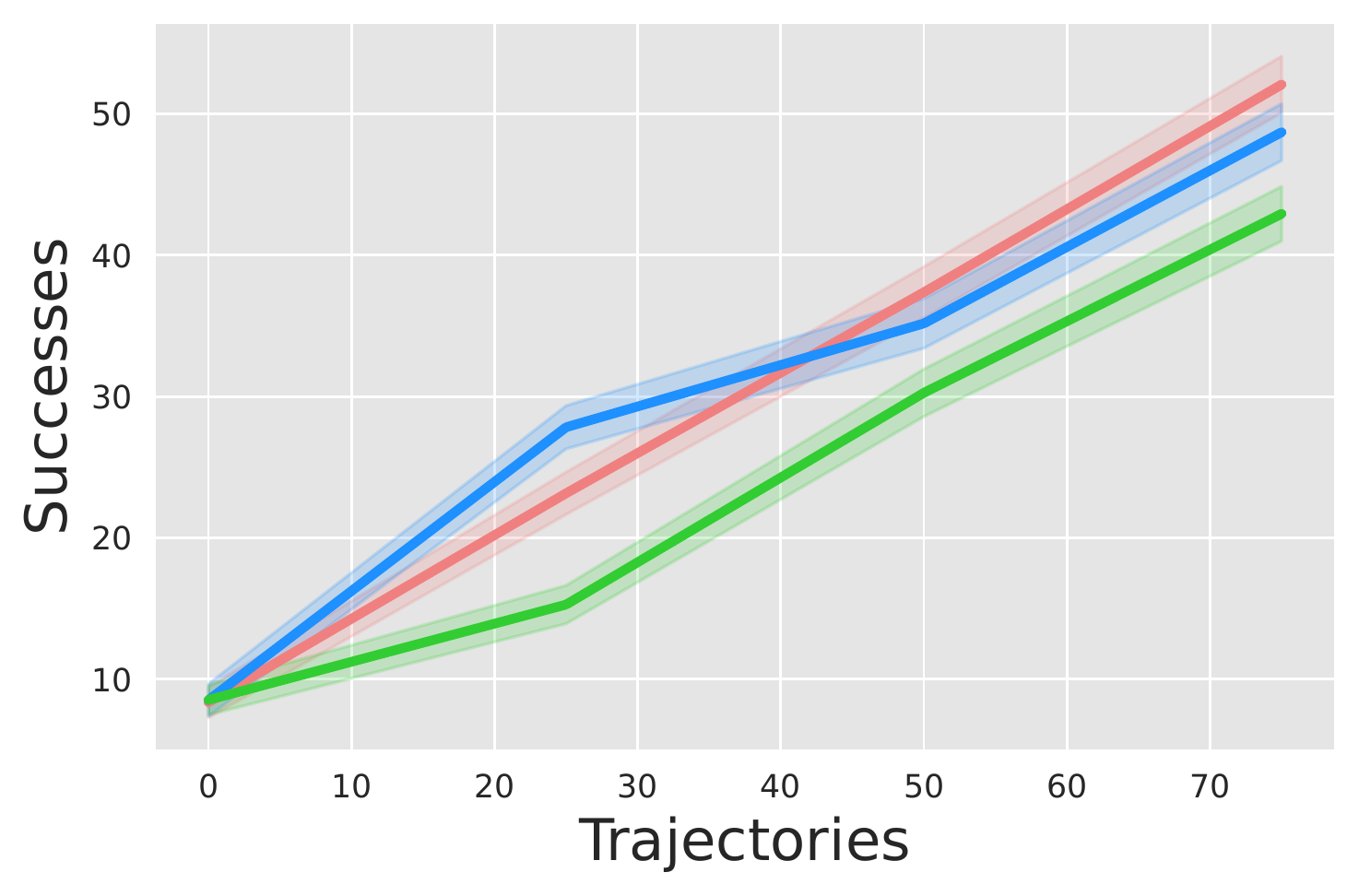}
   
    \caption{\small Stovepot}
    \end{subfigure}
   
    \caption{\small Coincidental success for exploration on our six tasks, where the robot reaches a semantically meaningful state while collecting data during exploration. We can see that \ours performs consistently well across tasks, and that just maximizing the change metric \texttt{AWR, EC} also yields much better data than previous state of the art approach \texttt{LEXA}. }
    \vspace{-0.25in}
    \label{fig:exploration-coinc-plots}
\end{figure*}

\section{Results}

\subsection{Exploration} 
We need a metric to evaluate the quality of the exploration data. While the change image norm is a good proxy for measuring object interaction, it does not consider if the different states are semantically interesting. Thus we define a metric that measures the \textit{number} of successful interactions, which are are determined by a human operator, as follows : 
\begin{itemize}
    \item Cabinet, fridge, shelf doors - has been opened or closed 
    \item Knife - lifted up
    \item Pan - unhooked, fully removed from hanger
    \item Pot - pushed/lifted/knocked over
\end{itemize}
Using this success criteria, we present evaluation of the exploratory data collected, in Figure~\ref{fig:exploration-coinc-plots}. For each task we run about 100-150 trajectories, and plot the cumulative number of successful exploration trajectories against the total number of trajectories seen during the exploration phase.

 We can see that \ours (red) outperforms or matches all other approaches in five out of six tasks, and also sees large number of successes for the top shelf. Further, we see that just maximizing the environment-change metric using \texttt{EC} or \texttt{AWR} leads to much better performance than LEXA, the previous state-of-the-art self-supervised exploration approach. We find that because the robot arm takes up a large portion of the observation, \texttt{LEXA} tries to collect data to resolve modelling inaccuracies of the arm. This is especially the case for tasks where random interactions are less likely to produce significant changes in the object, such as the particularly challenging knife task where \texttt{LEXA} never sees the picking up behavior. Further we see that on this task, having the agent-centric module which maximizes uncertainty in change space significantly improves performance over \texttt{EC} and \texttt{AWR}. For tasks like the top shelf which require less precise control, simply maximizing environment change is sufficient to collect high-quality data. However, even with slightly more involved control, such as the fridge task which requires the same object motion but has the robot in a more constrained position, addressing modelling inaccuracies in the change prediction is more critical. Moreover, using the agent-centric module leads to more robust performance for goal reaching, as described in the next section.

\subsection{Achieving Goals}

\begin{table}[t]
\begin{center}
\setlength{\tabcolsep}{5pt}
\begin{tabular}{lcccc}
\toprule
& \textbf{Cabinet} & \textbf{Knife} &  \textbf{Fridge} & \textbf{Top Shelf} \\
\midrule
\texttt{LEXA} \cite{mendonca2021discovering} & 0.20 & 0.00 & 0.00 & 0.00\\
\texttt{EC}   & 0.70 & 0.00 & 0.50 & \textbf{0.90} \\
\texttt{AWR} \cite{peng2019advantage} & 0.50 & 0.00 & - & - \\
\midrule
\texttt{ALAN} (ours) & \textbf{1.00}  & \textbf{0.60}  & \textbf{0.70} & 0.80\\
\bottomrule
\end{tabular}
\caption{\small Success rate for goal reaching. \ours is the only approach to get success on the challenging knife pick-up task, and just maximizing change (\texttt{EC}) is also much stronger than \texttt{LEXA}.}
\label{tab:ach}
\end{center}
\vspace{-0.3in}
\end{table}

Given the exploration data collected, can it be used to perform useful human specified tasks ? For this, we use the nearest-neighbor (kNN) approach outlined in section \ref{sec:achiever}, paired with model-based refinement to reach different human-specified goals. Specifically, once the kNN approach finds a trajectory, we use the action sequence as the mean of the initial sampling distribution of the CEM optimizer. The goals consist of a fully open fridge, cabinet or shelf, and a picked-up knife, as shown in Figure~\ref{fig:goal-imgs}. Since \texttt{AWR} has almost identical results for exploration and goal-reaching to \texttt{EC} on the first kitchen, and since they both optimize the same objective, we did not run it on the second kitchen (and therefore for the fridge and top shelf tasks). For each task, we run kNN on the exploratory data, in a visual feature space \cite{nair2022r3m} and select the best trajectory to execute conditioned on the start and goal images. We execute the top two trajectories five times each, collecting 10 different trials and present average success rates in Table~\ref{tab:ach}. We can see that our approach performs consistently well across tasks. Without the agent-centric module, there is no success on the difficult knife task, and overall performance across the remaining tasks is worse in terms of robustness. Moreover these results demonstrate the effectiveness of the environment change metric, since \texttt{LEXA} shows no success for three of the four tasks. 

\section{Discussion and Limitations}
\label{sec:conclusion}

We present \ours, an autonomously exploring agent that can efficiently explore in challenging real world environments. Our approach computes change in the environment, and utilizes it both directly as an environment-centric signal, as well as modelling the change and taking actions that maximize uncertainty in change space, which provides agent-centric signal. This reward in the absence of true task rewards helps our agent autonomously discover manipulation skills and perform useful tasks without any supervision. In the future, we hope to investigate distilling exploration data into a general goal-achieving policy, and studying continual learning across different tasks using a joint world model.




\section*{ACKNOWLEDGMENT}
This work was supported by DARPA Machine Common Sense, ONR MURI N00014-22-1-2773 and Sony Faculty Research Award.

\bibliographystyle{IEEEtran}
\bibliography{IEEEabrv,main}

\end{document}